\documentclass[conference]{IEEEtran}

\usepackage{cite}
\usepackage{amsmath,amssymb,amsfonts}
\usepackage{algorithmic}
\usepackage{graphicx}
\usepackage{textcomp}
\usepackage{xcolor}
\usepackage{booktabs}
\usepackage{multirow}
\usepackage{url}
\usepackage{balance}

\IEEEoverridecommandlockouts

\begin{document}

\title{A Picture is Worth a Thousand Tokens: How Vision Language Models Cut AI Energy Costs While Improving Accuracy}

\author{
\IEEEauthorblockN{Bhavika Jalli, Nikhil Korati Prasanna, and Jayanta Choudhury}
\IEEEauthorblockA{Ericsson\\
\{bhavika.jalli, nikhil.korati.prasanna, jayanta.choudhury\}@ericsson.com}
\thanks{Accepted at the 14th European Conference on Renewable Energy Systems (ECRES), July 7--9, 2026, London, UK.}
}

\maketitle

\begin{abstract}
LLM inference accounts for over 90\% of AI operational energy, scaling directly with input token count---a critical inefficiency for telecom network analytics and numerical time-series data analysis (NTSDA), where raw multivariate KPI windows from 4G/5G cell sites expand into thousands of floating-point tokens. Vision-Language Models (VLMs) eliminate this mismatch by encoding time-series as 2D plots, achieving 3.6--10.4x input token reduction across Llama-3.2-90B, Qwen2.5-VL-72B, and Pixtral-12B architectures. This translates to 1.8--2.5x measured inference energy reduction, saving approximately 7.2~MJ/day at telecom edge deployments and CloudRAN that monitor 200 cells per 15-minute interval. Critically, efficiency gains do not sacrifice accuracy: a fine-tuned Llama-3.2-90B-Vision VLM achieves 220.7\% higher precision than its text-only counterpart and outperforms LSTM and ARIMA baselines by over 144\% on telecom anomaly detection. On public benchmarks, Pixtral-12B achieves a 20.6x improvement in J/F1 score at mean F1 = 0.82. At 24 KPIs, text representations exceed the 128K context window of most production LLMs, rendering text-only processing infeasible, while visual representations remain within standard limits. These results establish VLMs as an energy-efficient and accuracy-superior modality for numerical time-series workloads, providing empirical grounding for AI inference systems that treat energy consumption as a first-class engineering constraint.
\end{abstract}


\section{Introduction}

US AI data-center power demand is projected to grow from 4~GW in 2024 to 123~GW by 2035 (30$\times$)~\cite{deloitte2025}, with data centers consuming 4.4\% of US electricity in 2023 and global AI demand potentially reaching 21\% by 2030~\cite{brookings2025}. Grid capacity is cited as the most critical challenge by 72\% of power executives, with interconnection queues stretching to seven years~\cite{deloitte2025}. Reducing per-inference energy is therefore an engineering prerequisite for sustainable AI deployment at scale.

Large Language Model (LLM) inference dominates AI energy consumption, accounting for over 90\% of lifecycle power~\cite{jegham2025}. Measurement studies confirm that inference energy scales linearly with input token count~\cite{caravaca2025}, establishing token count as the primary controllable lever for energy reduction. Unlike one-time training, inference runs continuously: even modest per-query savings compound across millions of daily calls.

This token-energy coupling is particularly wasteful for numerical time-series data analysis (NTSDA). When LLMs ingest raw time-series, a modest multivariate KPI window explodes into tens of thousands of floating-point tokens despite carrying far less information density than natural language. Tokenization adds no representational value to numerical sequences and introduces error risk from improper token patterns~\cite{singh2024} or mismatched tokenization models~\cite{mostafa2025}. A single 8-KPI window of 381 time points generates 46,000--60,000 tokens, exceeding context windows of production models and approaching memory ceilings of edge GPUs, making text-based analysis both energy-wasteful and often physically infeasible.

Vision-Language Models (VLMs) eliminate this mismatch by rendering time series as 2-D plots, compressing temporal information into far fewer visual tokens. In our telecom anomaly detection pipeline, a fine-tuned VLM achieves 220.7\% higher precision and 2.5x energy savings over its text-only counterpart, while outperforming LSTM and ARIMA baselines by over 144\%. On public benchmarks, a 20.6x improvement in J/F1 efficiency was achieved~\cite{he2025}. This paper makes the following contributions:
\begin{enumerate}
\item The first direct energy comparison of LLM vs.\ VLM inference for time-series anomaly detection across three vision encoder architectures.
\item Empirical measurements on public benchmarks and live telecom data demonstrating 1.8--2.5x energy savings with simultaneous accuracy improvements.
\item Analysis of image compression as an additional energy reduction lever.
\item Architecture-specific guidance for energy-constrained edge deployments.
\item Operational scaling implications for production telecom environments.
\end{enumerate}

\section{Related Work}

Input token count carries a substantial energy penalty: increasing input length from 100 to 900 tokens at fixed 100-token output raises energy by 2.19x~\cite{caravaca2025}, confirming prefill cost as a significant and independently actionable efficiency target. LLM inference energy scales directly with prompt length; longer contexts increase joules-to-first-token and reduce throughput by saturating the GPU~\cite{niu2025}. This token-energy coupling makes LLMs inherently inefficient for numerical time-series data, where thousands of floating-point tokens encode information that a compact visual representation conveys at a fraction of the energy cost.

VLMs incur higher per-token power draw than LLMs due to visual encoder overhead~\cite{chung2025}, imposing significant computational and energy demands~\cite{kalzhan2025}. However, visual tokens carry substantial redundancy; only a small fraction is essential for accurate response generation~\cite{he2026}. VLMs consistently outperform LLM-based time-series anomaly detection while consuming up to 36x fewer tokens per variable~\cite{he2025}. Visual representations amplify coarse-grained temporal structures that raw numerical tokens fail to capture, improving both range-wise and variate-wise anomaly localization~\cite{liu2024}. With few-shot visual prompting alone, VLMs surpass all supervised baselines on deterministic reasoning tasks, delivering up to 433\% performance improvement over numerical approaches~\cite{liu2024}.

\textbf{Edge Deployment Constraints.} The energy efficiency argument becomes a hard feasibility constraint at telecom edge sites. Shi et al.~\cite{shi2026} profile LLM fine-tuning on a single NVIDIA RTX A6000 (48~GB VRAM), representative of O-RAN near-RT RIC and MEC accelerators, and demonstrate that sequence lengths exceeding 60,000 tokens trigger out-of-memory errors on Qwen architectures under 4-bit quantization, with 50,000 tokens at 0.7 GPU utilization established as the safe operating ceiling. The text-only representation of an 8-KPI time-series window in our experiments requires 46,101 to 59,803 tokens depending on the tokenizer, approaching or exceeding this hardware limit. In contrast, the VLM visual modality compresses the same signal into 5,442 to 16,800 vision tokens, remaining well within the edge GPU's operating envelope. For single-GPU edge deployments, the VLM token reduction is therefore not merely an efficiency gain but a deployment prerequisite. At the cluster level, Kang et al.~\cite{kang2022} demonstrate that energy-aware scheduling of heterogeneous GPU deep learning jobs reduces operational energy costs by exploiting time-varying electricity pricing and dynamic right-sizing, reinforcing the case for per-query energy measurement at the model level.

To our knowledge, no existing study directly compares the end-to-end energy consumption of LLMs and VLMs for time-series anomaly detection. While VLMs incur higher per-token energy costs from visual encoder overhead and modality-fusion layers, they achieve an order-of-magnitude reduction in total token count by compressing numerical sequences into compact 2D plot representations, a ratio that scales with input window length. This trade-off motivates the central empirical question: whether VLM token compression yields a net inference energy reduction that outweighs the higher per-token cost, establishing VLMs as the more energy-efficient paradigm for numerical time-series workloads.

\section{Methodology}

\begin{figure}[t]
\centering
\includegraphics[width=\columnwidth]{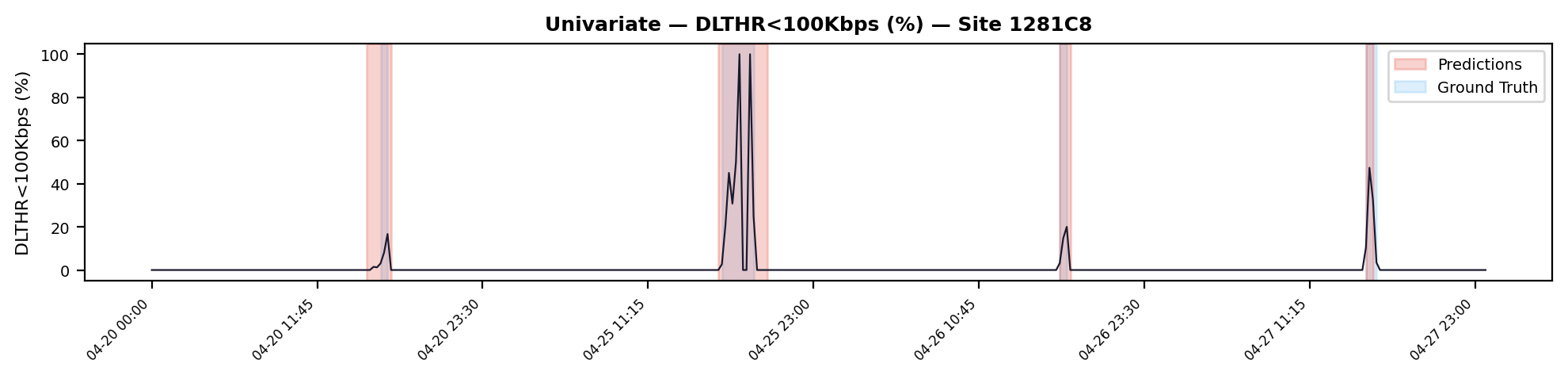}
\caption{Univariate time-series rendering of an example KPI, DLTHR$<$100Kbps (\%), for a single cell site. Red shading indicates VLM-detected anomaly intervals; blue shading indicates ground-truth labels.}
\label{fig:univariate}
\end{figure}

The first stage renders 2D graphs of univariate time-series (Fig.~\ref{fig:univariate}) as raster images, mapping time to the horizontal axis and measurement values to the vertical axis. Both axes are stripped of tick marks, numerical labels, and gridlines, following VLM4TS~\cite{he2025}, to eliminate textual artifacts from the pixel space.

\begin{figure}[t]
\centering
\includegraphics[width=\columnwidth]{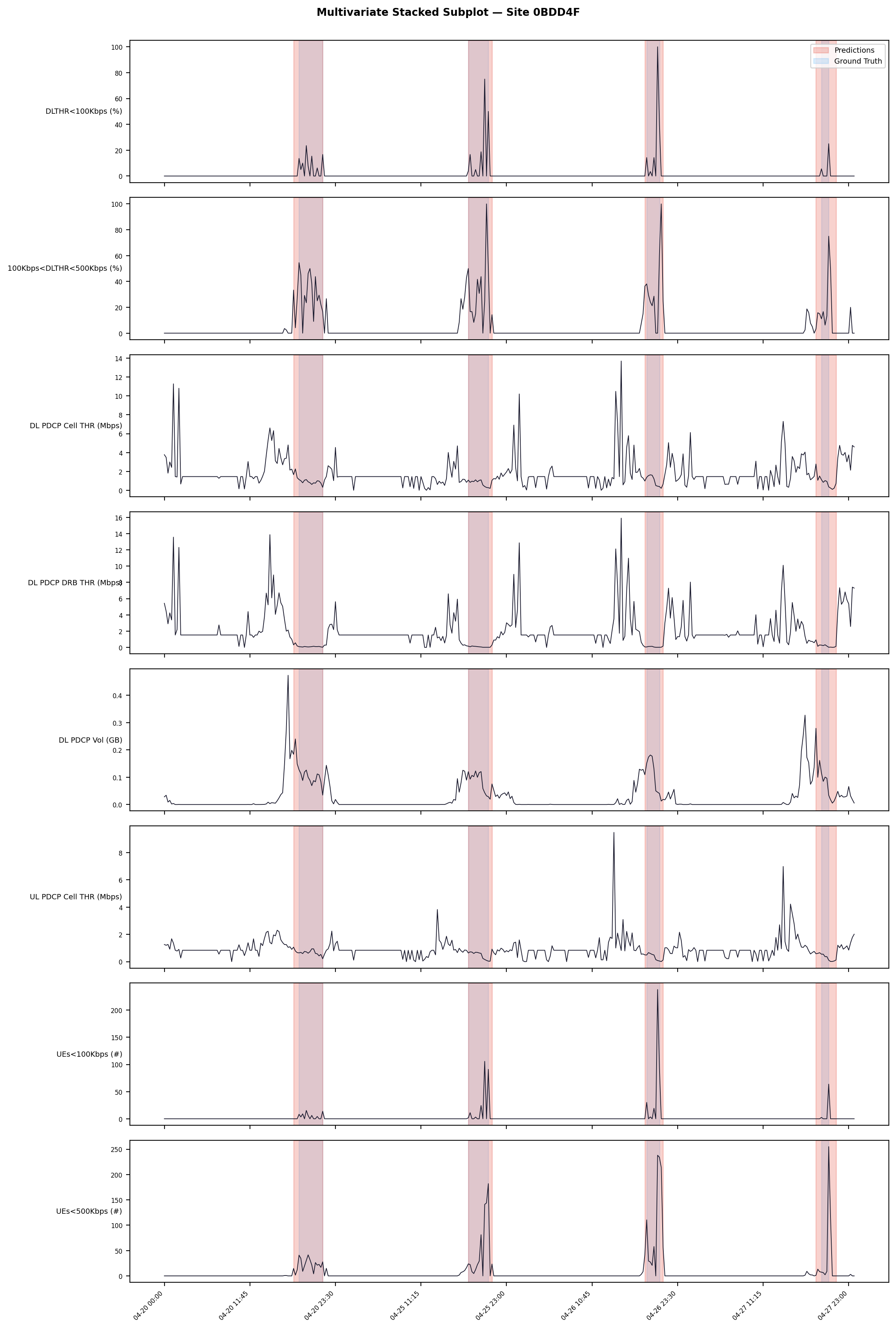}
\caption{Multivariate stacked subplot rendering of 8 Throughput KPIs for a single cell site. Each subplot shares a common temporal axis. Red shading indicates VLM-detected anomaly intervals; blue shading indicates ground-truth labels.}
\label{fig:multivariate}
\end{figure}

The next stage performs patch generation and anomaly interval proposal following the VLM4TS~\cite{he2025} temporal segmentation procedure. Our approach extends VLM4TS beyond single-panel univariate graphs to vertically stacked subplot compositions (Fig.~\ref{fig:multivariate}), where each subplot renders a distinct univariate component of a multivariate time-series. Subplots share a common temporal axis, preserving inter-variable alignment while maintaining per-variable visual separation.

The stacked-subplot formulation is geometrically motivated: a multivariate time-series of dimension $d$ traces a trajectory on a $d$-dimensional manifold, and a vertically aligned stack of $d$ univariate subplots constitutes a visual cross-section of that manifold. Each subplot contributes one coordinate of the instantaneous state vector, enabling the vision model to jointly attend to co-occurring morphological features across variables.

Next, structured prompts condition VLM inference over rendered images, encoding the detection objective alongside domain-specific morphological primitives (spike, hump, oscillatory deviation) and their geometric attributes to steer attention toward waveform geometry.

\subsection{Token Count Comparison}

He et al.~\cite{he2025} reported a $\sim$30$\times$ reduction in input tokens when time-series data is presented as a rendered image, measured on single-panel univariate plots. Our stacked-subplot encoding of multivariate data produces composite images, whose pixel dimensions scale linearly with the number of variables, increasing the vision encoder's token budget proportionally. We therefore reassess whether the token-efficiency advantage of He et al.\ holds---and to what degree---as image resolution grows with signal dimensionality. Precise token accounting also provides the denominator for per-token energy metrics. Since inference energy scales with token throughput~\cite{he2025}, characterizing the token budget per modality enables normalized cost comparison.

\subsection{Energy Measurement}

GPU energy is measured using the Zeus~\cite{zeus2023} framework, reading cumulative hardware energy counters via NVML (\texttt{nvmlDeviceGetTotalEnergyConsumption}) to avoid sampling error from power-polling approaches. As a secondary source, instantaneous power draw is polled via NVML at 50~ms intervals and integrated using the trapezoidal rule. For multi-GPU deployments, energy is summed across all devices. Each configuration (model $\times$ modality) is evaluated over 3 independent runs of 256 output tokens, preceded by a warmup pass to stabilize GPU clock frequencies.

\subsection{Model Selection}

Three VLMs are selected across principal vision encoder strategies, ensuring that findings are not artifacts of a single tokenization scheme. Llama-3.2-90B-Vision~\cite{llama2024} uses cross-attention fusion with a fixed 6,404-token visual representation regardless of resolution. Qwen2.5-VL-72B~\cite{qwen2024} applies dynamic-resolution patching, scaling the token budget with image height and number of KPI subplots. Pixtral-12B~\cite{pixtral2024} decomposes images into 16$\times$16 pixel patches, yielding the highest vision token counts, while preserving spatial detail for narrow waveform feature detection.

\section{Experimental Setup}

For public data, all models are loaded with 4-bit NF4 quantization on three NVIDIA RTX A6000 GPUs (48~GB each). Each signal is evaluated under text modality with raw time-series serialized as comma-separated floats and tokenized natively, and image modality. Inference energy is measured per query via NVML hardware counters at 50~ms polling intervals. VLM4TS Stage~1 (CLIP-based visual screening for candidate anomaly window identification) is included as a pipeline baseline. For telecom data, deployment configurations reflect realistic operational constraints: Llama-3.2-90B across 4$\times$ A100 GPUs (4-bit); Qwen2.5-VL-72B across 2$\times$ A100 GPUs (4-bit); Pixtral-12B on a single A6000 (bfloat16). Token counts are measured using each model's native processor: Llama via text tokenizer and fixed 6,404-token cross-attention vision projection; Qwen2.5-VL-72B via AutoProcessor with dynamic-resolution patching; Pixtral via LlavaProcessor with 16$\times$16 patch tokenization.

\subsection{Public Dataset}

We evaluate the realAWSCloudwatch, a subset of the VLM4TS benchmark~\cite{he2025}, comprising univariate time series from cloud infrastructure monitoring. We select 17 candidate signals; 10 are retained per model, stratified by series length (1,243--4,730 time points). Ground truth anomaly intervals are sourced from the Sintel Orion benchmark~\cite{he2025}. Detection performance is reported as F1 score at a significance threshold $\alpha = 0.01$.

\subsection{Telecom Specific Experiment}

Experiments use operational KPI data from a live 4G/5G telecom network, spanning 9 dates and 209 cells in April 2025: 5 training dates (April 11--13, 18--19) and 4 test dates (April 20, 25--27). Each cell reports 24 KPIs at 15-minute intervals, 96 time points per day, and $\sim$381 per cell across the test period. Ground-truth anomaly labels are derived from a Causal Anomaly Detection system~\cite{shi2025causal}. For evaluation, we evaluate only for dates with ground-truth anomalies, avoiding a penalty for anomaly-free days. For the text modality, all 8 KPI time-series were serialized as comma-separated floating-point values (one row per time step) and tokenized using each model's native tokenizer. For the image modality, the same 8 KPIs were rendered as a stacked subplot chart (one subplot per KPI, shared time axis) at 150~DPI and processed through each model's vision encoder to produce vision tokens.

\section{Results and Analysis}

\subsection{Public Dataset Results and Analysis}

\begin{figure}[t]
\centering
\includegraphics[width=\columnwidth]{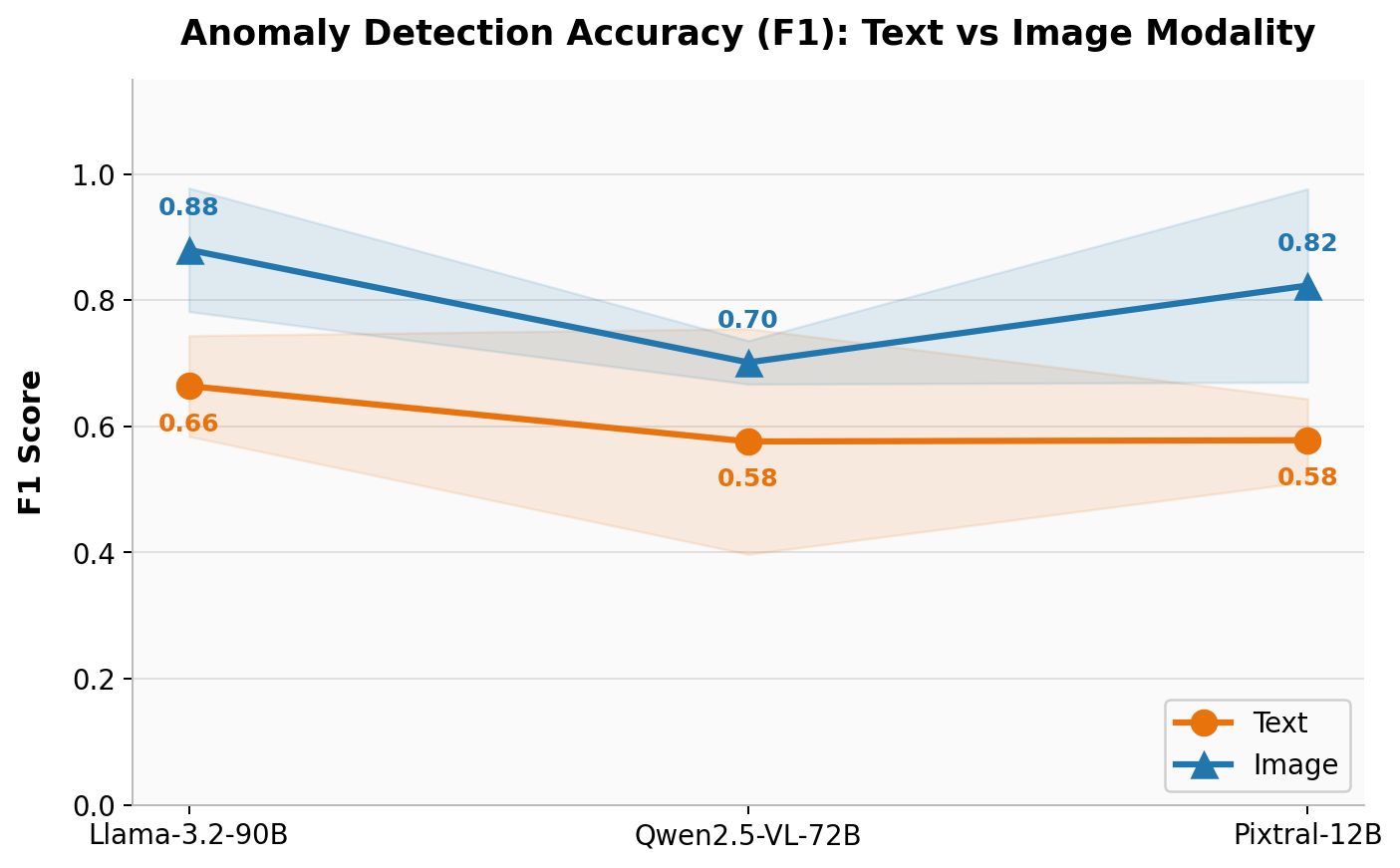}
\caption{Mean F1 score for time-series anomaly detection on the realAWSCloudwatch benchmark across three VLM architectures under text-only (orange) and image (blue) input modalities.}
\label{fig:f1_public}
\end{figure}

\begin{figure}[t]
\centering
\includegraphics[width=\columnwidth]{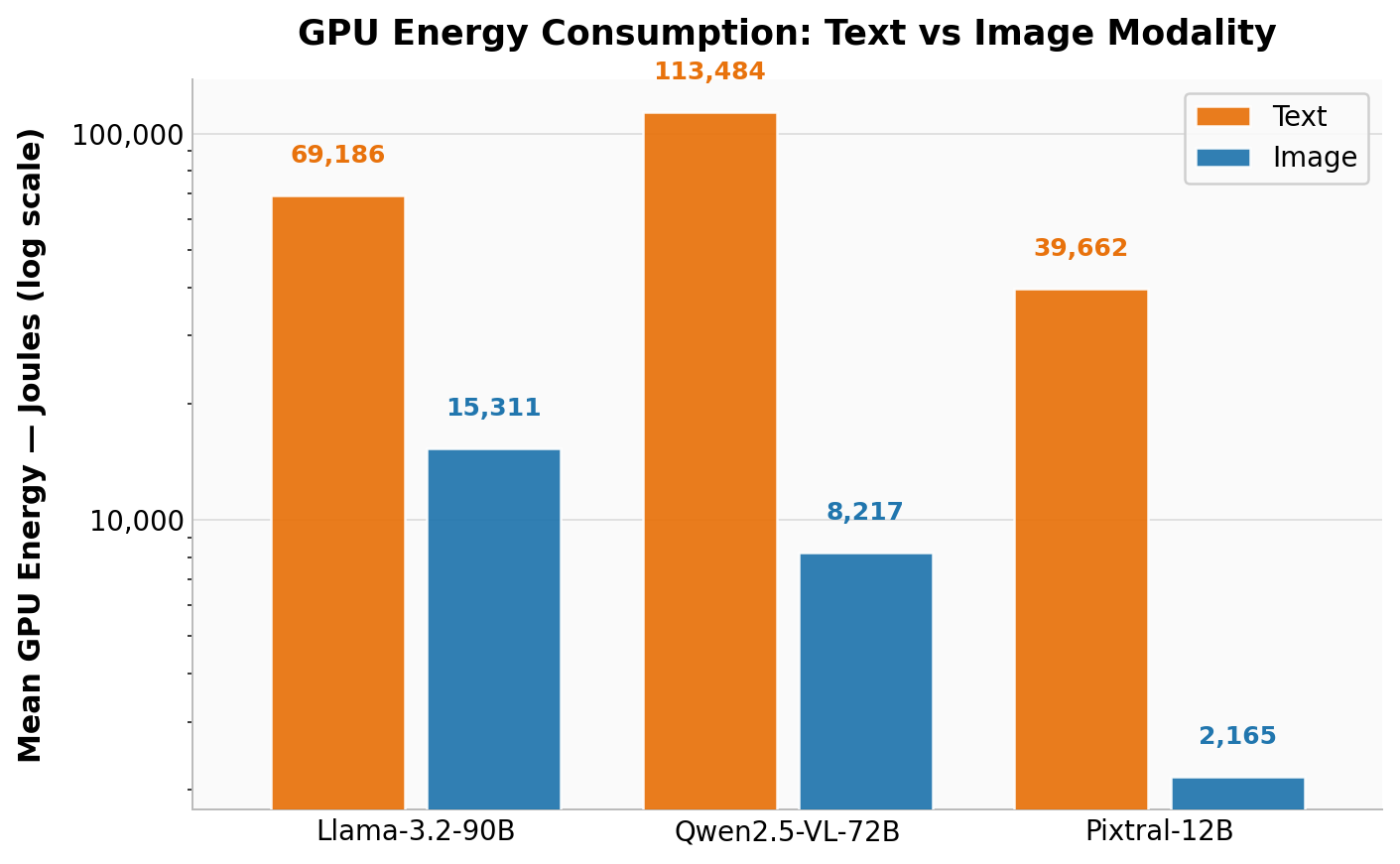}
\caption{Mean GPU energy consumption per inference query (Joules, log scale) for text-only and image modalities across three VLM architectures on the realAWSCloudwatch benchmark.}
\label{fig:energy_public}
\end{figure}

Figures~\ref{fig:f1_public} and~\ref{fig:energy_public} report F1 scores ($\pm$1 std.\ dev.\ by shaded bands across five signals) and GPU energy consumption, respectively. Image modality consistently outperforms text across all architectures (mean F1: 0.70--0.88 vs.\ 0.58--0.66). Energy, measured via NVML hardware counters on three NVIDIA RTX A6000 GPUs, is reduced by image modality relative to text by 4.5$\times$ (Llama-3.2-90B), 13.8$\times$ (Qwen2.5-VL-72B), and 18.3$\times$ (Pixtral-12B). Table~\ref{tab:energy_public} details the per-query energy, inference time, and average power draw for each model-modality configuration.

\begin{table}[t]
\centering
\caption{Measured inference energy consumption per query on the realAWSCloudwatch dataset (NVIDIA RTX A6000, 4-bit NF4 quantization)}
\label{tab:energy_public}
\begin{tabular}{lcrrc}
\toprule
\textbf{Model} & \textbf{Modality} & \textbf{Energy (J)} & \textbf{Time (s)} & \textbf{Red.} \\
\midrule
Llama-3.2-90B & Text-only & 69,187 & 168.4 & -- \\
Llama-3.2-90B & VLM image & 15,311 & 37.2 & 4.5x \\
\midrule
Qwen2.5-VL-72B & Text-only & 113,484 & 279.2 & -- \\
Qwen2.5-VL-72B & VLM image & 8,218 & 20.8 & 13.8x \\
\midrule
Pixtral-12B & Text-only & 39,663 & 99.7 & -- \\
Pixtral-12B & VLM image & 2,166 & 7.1 & 18.3x \\
\bottomrule
\end{tabular}
\end{table}

\begin{table}[t]
\centering
\caption{F1/Joule efficiency metric across models and modalities on the realAWSCloudwatch dataset}
\label{tab:jf1}
\begin{tabular}{lccc}
\toprule
\textbf{Model} & \textbf{Text J/F1} & \textbf{Image J/F1} & \textbf{Impr.} \\
\midrule
Llama-3.2-90B & 84,034 & 16,756 & 5$\times$ \\
Qwen2.5-VL-72B & 197,617 & 10,954 & 18$\times$ \\
Pixtral-12B & 52,356 & 2,538 & 20$\times$ \\
\bottomrule
\end{tabular}
\end{table}

\textbf{Energy-Accuracy Efficiency Analysis.} Table~\ref{tab:jf1} reports J/F1 (GPU energy per unit F1 score, averaged across five signals with F1 = 0 excluded; lower is better). Pixtral-12B achieves the best image efficiency at 2,538~J/F1 versus 52,356~J/F1 for text (20.6x gain), delivering mean F1 = 0.82 at the lowest absolute energy (2,166~J/query). Qwen2.5-VL-72B text is the least efficient at 197,617~J/F1, making it operationally impractical for high-frequency monitoring. Llama's smaller gap reflects its compact tokenizer, which keeps text token counts lower than Qwen and Pixtral. The J/F1 metric provides a hardware-grounded criterion for modality selection in energy-constrained deployments.

\subsection{Telecom Dataset Results and Analysis}

\begin{table}[t]
\centering
\caption{Input token counts for text-only vs.\ visual modality (8 KPIs, 381 time points) on the Telecom dataset}
\label{tab:tokens}
\begin{tabular}{lccc}
\toprule
\textbf{Model} & \textbf{Text Tokens} & \textbf{Vision Tokens} & \textbf{Red.} \\
\midrule
Llama-3.2-90B-Vision & 46,101 & 6,404 & 7.2x \\
Qwen2.5-VL-72B & 56,372 & 5,442 & 10.4x \\
Pixtral-12B & 59,803 & 16,800 & 3.6x \\
\bottomrule
\end{tabular}
\end{table}

Table~\ref{tab:tokens} summarizes input token counts. Reductions are architecture-dependent: Llama's cross-attention fusion projects any image into a fixed 6,404-token budget (7.2$\times$ reduction), Qwen's dynamic resolution tiles images into 28$\times$28 patches with adjacent-token merging for the largest compression at 10.4$\times$, while Pixtral's 16$\times$16 patch tokenization preserves spatial fidelity at a more modest 3.6$\times$. The variation in text token counts (46,101--59,803) reflects differing tokenizer efficiency for numerical data: Qwen and Pixtral split floating-point numbers into more sub-word tokens than Llama, making the visual pathway even more advantageous for these architectures. Since inference energy scales linearly with input token count~\cite{jegham2025,niu2025}, these reductions translate directly to proportional prefill-phase energy savings.

\subsection{Anomaly Detection Accuracy}

\begin{table}[t]
\centering
\caption{Anomaly detection performance on telecom KPI data (per-day temporal evaluation) with Llama-3.2-90B}
\label{tab:accuracy}
\begin{tabular}{lcccc}
\toprule
\textbf{Method} & \textbf{P} & \textbf{R} & \textbf{F1} & \textbf{Tokens} \\
\midrule
ARIMA & 0.202 & 0.268 & 0.186 & 42,565 \\
LSTM & 0.190 & 0.293 & 0.190 & 42,565 \\
LLM text-only & 0.145 & 0.415 & 0.185 & 42,565 \\
0-shot VLM & 0.298 & 0.738 & 0.360 & 6,404 \\
SFT VLM & 0.465 & 0.647 & 0.464 & 6,404 \\
\bottomrule
\end{tabular}
\end{table}

Table~\ref{tab:accuracy} compares the fine-tuned Llama-3.2-90B-Vision against baselines. The SFT VLM achieves precision = 0.465, a 220.7\% improvement over the text-only LLM (0.145), while consuming 7.2x fewer tokens. Visual rendering preserves temporal morphologies as spatially coherent patterns, whereas text serialization fragments them across tokenization boundaries. The zero-shot VLM already outperforms all text-based methods (F1 = 0.360 vs.\ 0.185--0.190), confirming the visual representation itself provides a fundamental advantage. Since LoRA fine-tuning modifies only attention weights, token count and energy cost remain identical for zero-shot and fine-tuned VLMs.

\subsection{Token Scaling with KPI Count}

\begin{figure}[t]
\centering
\includegraphics[width=\columnwidth]{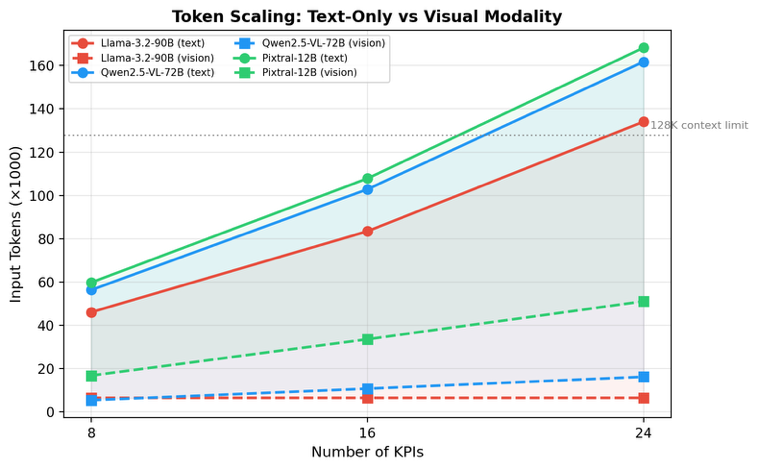}
\caption{Input token count scaling with KPI dimensionality for text-only (solid) and visual (dashed) modalities.}
\label{fig:token_scaling}
\end{figure}

Vision token growth with KPI count is architecture-dependent (Fig.~\ref{fig:token_scaling}): Qwen's dynamic tiling scales most aggressively with image height, while Llama's fixed-budget encoding remains constant regardless of subplot count. At 24 KPIs, all three text representations exceed 128K tokens ($\sim$2.9x from 8 to 24 KPIs) for all three models, surpassing the context window of most production deployments and forcing lossy truncation that the visual modality avoids entirely.

\subsection{Inference Energy Consumption}

\begin{table}[t]
\centering
\caption{Measured inference energy consumption per query (256 output tokens, NVIDIA A100/A6000)}
\label{tab:energy_telecom}
\begin{tabular}{lcrrr}
\toprule
\textbf{Model} & \textbf{Modality} & \textbf{Energy (J)} & \textbf{Time (s)} & \textbf{Power (W)} \\
\midrule
Llama-3.2-90B & Text-only & 59,751 & 113.2 & 528 \\
Llama-3.2-90B & VLM image & 23,557 & 46.3 & 509 \\
\midrule
Qwen2.5-VL-72B & Text-only & 4,363 & 17.8 & 245 \\
Qwen2.5-VL-72B & VLM image & 2,397 & 13.2 & 182 \\
\midrule
Pixtral-12B & Text-only & 7,768 & 26.1 & 297 \\
Pixtral-12B & VLM image & 3,151 & 10.8 & 291 \\
\bottomrule
\end{tabular}
\end{table}

Table~\ref{tab:energy_telecom} reports measured energy reductions of 2.5x on Llama (59,751~J vs.\ 23,557~J), 1.8x on Qwen (4,363~J vs.\ 2,397~J), and 2.5x on Pixtral (7,768~J vs.\ 3,151~J). The gap narrows relative to token savings because the visual encoder imposes a constant overhead per query, but the net balance decisively favors image modality at all scales tested. At operational scale (209 cells queried per 15-minute interval), the 2.5x per-query reduction compounds to $\sim$7.2~MJ/day savings per model. Text inputs were truncated to each model's maximum context length (40K for Llama, 32K for Qwen/Pixtral); the full 8-KPI time series would require chunking or a longer-context model, further increasing text-only costs.

\subsection{Image Compression and Energy Efficiency}

To investigate whether further energy savings can be achieved by reducing image resolution, we evaluated the effect of image compression on vision token count, inference energy, and detection accuracy. Using Qwen2.5-VL-72B in zero-shot mode with ontology-guided proposals, we rendered the same 8-KPI stacked subplot charts at four DPI levels (150, 100, 75, and 50) in PNG format, plus a JPEG-85 variant at 75~DPI. All 108 cells with ground-truth anomalies were evaluated under each configuration.

\begin{table}[t]
\centering
\caption{Effect of image compression on vision tokens, inference energy, and anomaly detection accuracy (Qwen2.5-VL-72B, zero-shot, 108 cells)}
\label{tab:compression}
\begin{tabular}{lccc}
\toprule
\textbf{Config} & \textbf{Vision Tokens} & \textbf{Energy (J)} & \textbf{F1} \\
\midrule
150 DPI PNG & 17,303 & 8,818 & 0.347 \\
100 DPI PNG & 8,177 & 7,193 & 0.347 \\
75 DPI PNG & 5,112 & 6,660 & 0.354 \\
50 DPI PNG & 2,749 & 6,283 & 0.340 \\
75 DPI JPEG-85 & 5,112 & 6,727 & 0.351 \\
\bottomrule
\end{tabular}
\end{table}

Table~\ref{tab:compression} shows that reducing resolution from 150~DPI to 75~DPI cuts vision tokens by 70\% (17,303 to 5,112) and inference energy by 24\% (8,818~J to 6,660~J) with no degradation in detection accuracy (F1 = 0.354 vs.\ 0.347). Even at 50~DPI, where the image is reduced to approximately one-ninth of its original pixel area, F1 remains within 2\% of the baseline (0.340 vs.\ 0.347). JPEG compression at 75~DPI produces identical token counts to PNG at the same resolution, with negligible impact on accuracy (F1 = 0.351).

The stability of F1 across compression levels indicates that the coarse temporal morphologies relevant to anomaly detection, for example spikes, humps, and oscillatory deviations, are preserved even at low resolution. The energy savings are proportionally smaller than the token reduction because the output generation phase (256 tokens) constitutes a fixed cost that dominates at lower input counts. Nevertheless, the 24\% energy reduction from 150 to 75~DPI represents a practical optimization for edge deployments where image rendering resolution can be tuned without retraining the model.

\section{Discussion}

\subsection{Cross-Dataset Synthesis}

The consistent energy reduction (1.8--2.5x) and accuracy improvement (mean F1 image 0.72 vs.\ text 0.49 on public data; F1 0.464 vs.\ 0.185 on telecom) across architectures on both public (zero-shot, univariate) and telecom (fine-tuned, multivariate) datasets confirms that the efficiency gain is structural, arising from token compression, rather than an artifact of any single vision encoder design or domain-specific tuning.

\subsection{Architecture Selection Trade-offs}

Llama's fixed 6,404-token visual budget provides deterministic energy consumption amenable to capacity planning, but sacrifices spatial fidelity. Qwen's dynamic tiling achieves the strongest compression (10.4x) on standard inputs but scales aggressively with image height. Pixtral offers the most faithful spatial encoding at modest compression (3.6x). The choice depends on whether the deployment prioritizes energy predictability (Llama), maximum compression (Qwen), or spatial fidelity (Pixtral).

\subsection{Operational Energy Impact}

A telecom operator monitoring 209 cells at 15-minute intervals generates $\sim$20,064 queries/day. The VLM modality saves $\sim$7.2~MJ/day, equivalent to the daily residential energy consumption of 2.4 average US households. Extrapolating to network-wide deployment spanning thousands of cells and multiple KPI dimensions, cumulative savings reach hundreds of MJ daily. For edge deployments with limited thermal dissipation capacity, the 2.5x energy reduction translates directly to either higher query throughput within the same thermal envelope or reduced cooling infrastructure requirements.

\subsection{Limitations and Future Work}

All measurements use single-GPU configurations (NVIDIA A6000 and A100); multi-GPU parallel inference introduces communication overhead and memory transfer costs that may alter the token-energy relationship. Only three VLM architectures are evaluated; emerging designs such as mixture-of-experts vision models or sparse attention VLMs may exhibit different energy scaling properties. The fixed 256 output token constraint bounds generalizability to short-response tasks; production deployments generating longer diagnostic outputs would shift the energy balance toward output-dominated regimes. The telecom evaluation covers a single operator and market; cross-operator validation remains open. Future work should extend to full-system energy accounting, adaptive resolution strategies based on signal complexity, streaming inference with energy-aware batching, and carbon intensity calibration for sustainability reporting.

\section{Conclusion}

Our evaluation confirms that VLMs reduce inference energy for numerical time-series workloads while improving detection accuracy. Across three vision encoder architectures, the image modality achieves 3.6--10.4x token reduction and 1.8--2.5x energy savings, with Pixtral-12B delivering 20.6x improvement in energy-normalized efficiency (2,538 vs.\ 52,356~J/F1). The fine-tuned Llama-3.2-90B-Vision achieves 220.7\% higher precision than its text counterpart at 2.5x lower energy, outperforming LSTM and ARIMA baselines by over 144\%.

At 24 KPIs, text representations exceed the 128K context window of production LLMs, rendering text-only inference infeasible, while visual representations remain within standard limits. Resolution reduction to 75~DPI preserves detection accuracy while delivering an additional 24\% energy saving. At edge scale, the VLM pathway saves $\sim$7.2~MJ/day per model. For telecom edge sites constrained to single-GPU accelerators, the VLM token reduction is not merely an efficiency gain but a deployment prerequisite, as text-mode token counts exceed the demonstrated memory ceiling of representative edge hardware~\cite{shi2026}.

\balance
\bibliographystyle{IEEEtran}

\end{document}